%% file: eccv2022submission.tex
\documentclass[runningheads]{llncs}
\usepackage{graphicx}
\usepackage{booktabs}
\usepackage{multirow}
\usepackage{makecell}
\usepackage{float}
\usepackage{comment}
\usepackage{xspace}
\usepackage{wrapfig}
\usepackage{bm}
\usepackage{caption}
\usepackage{subcaption}
\usepackage{tikz}
\usepackage{comment}
\usepackage{amsmath} % define this before the line numbering.
\usepackage{amssymb}
\usepackage[normalem]{ulem}
\usepackage{color}
\usepackage{bm}

\usepackage[accsupp]{axessibility}  % Improves PDF readability for those with disabilities.

\usepackage[font=small,skip=8pt]{caption}
\input{customsetup}

\begin{document}
% \linenumbers
\pagestyle{headings}
\mainmatter
\def\ECCVSubNumber{5622}  % Insert your submission number here

\title{DiffuStereo: High Quality Human Reconstruction via Diffusion-based Stereo Using Sparse Cameras} % Replace with your title
% \title{FITE: Learning Clothed Humans using a First-Implicit-Then-Explicit Hybrid Representation} % Replace with your title

\begin{comment}
\titlerunning{ECCV-22 submission ID \ECCVSubNumber} 
\authorrunning{ECCV-22 submission ID \ECCVSubNumber} 
\author{Anonymous ECCV submission}
\institute{Paper ID \ECCVSubNumber}
\end{comment}
%****************** \email{linsy21@mails.tsinghua.edu.cn}}
% \end{comment}
% \begin{comment}
\titlerunning{DiffuStereo}
\author{Ruizhi Shao \and
Zerong Zheng \and
Hongwen Zhang \and
Jingxiang Sun \and \\
Yebin Liu}
\authorrunning{R. Shao et al.}
% \institute{Tsinghua University, Princeton NJ 08544, USA \and
% \end{comment}
\institute{Tsinghua University, Beijing, China}%\\
%******************
\maketitle

\begin{abstract}
We propose DiffuStereo, a novel system using only sparse cameras (8 in this work) for high-quality 3D human reconstruction. At its core is a novel diffusion-based stereo module, which introduces diffusion models, a type of powerful generative models, into the iterative stereo matching network. To this end, we design a new diffusion kernel and additional stereo constraints to facilitate stereo matching and depth estimation in the network. We further present a multi-level stereo network architecture to  handle high-resolution (up to 4k) inputs without requiring unaffordable memory footprint. Given a set of sparse-view color images of a human, the proposed multi-level diffusion-based stereo network can produce highly accurate depth maps, which are then converted into a high-quality 3D human model through an efficient multi-view fusion strategy. Overall, our method enables automatic reconstruction of human models with quality on par to high-end dense-view camera rigs, and this is achieved using a much more light-weight hardware setup. Experiments show that our method outperforms state-of-the-art methods by a large margin both qualitatively and quantitatively.

\keywords{3D Human Reconstruction, Diffusion Model, Multiview}
\end{abstract}

\input{section_intro}
\input{section_relwork}
\input{section_method}
\input{section_exps}

\input{section_conclusion}

\bibliographystyle{splncs04}
\bibliography{egbib}
\end{document}

%% file: customsetup.tex
\usepackage[capitalize]{cleveref}
\crefname{section}{Sec.}{Secs.}
\Crefname{section}{Section}{Sections}
\Crefname{table}{Table}{Tables}
\crefname{table}{Tab.}{Tabs.}

\makeatletter
\DeclareRobustCommand\onedot{\futurelet\@let@token\@onedot}
\def\@onedot{\ifx\@let@token.\else.\null\fi\xspace}

\def\ie{{i.e}\onedot}

\makeatother

%% file: section_intro.tex
\section{Introduction}
\begin{figure}[t]
    \centering
    \includegraphics[width=\linewidth]{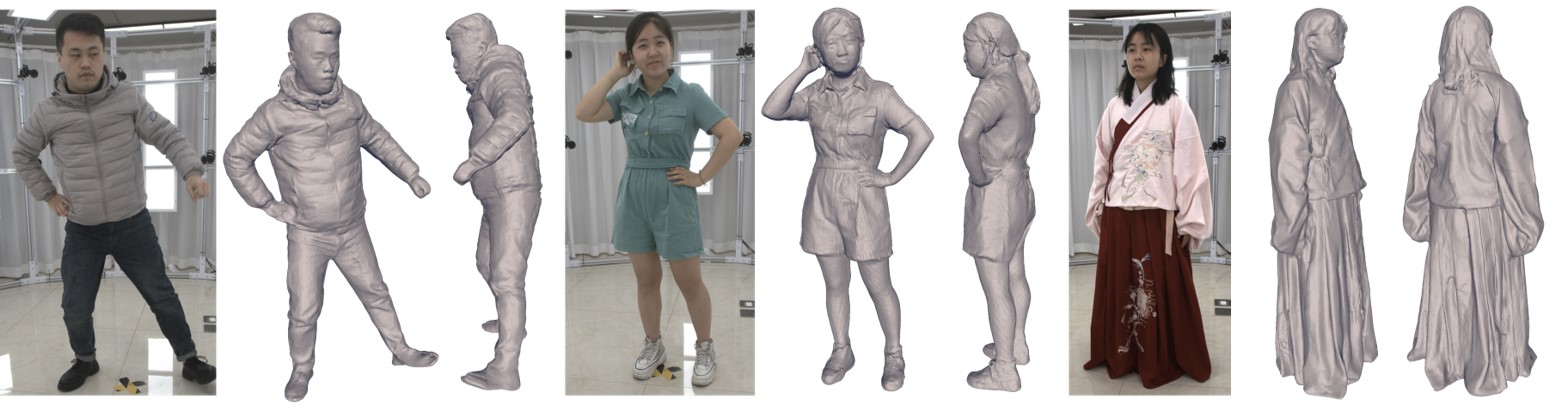}
    \caption{Our DiffuStereo system can reconstruct high accurate 3D human models with sharp geometric details using only 8 RGB cameras. Such results can only be achieved using nearly hundred of cameras before. See Supp. video for more video results. }
    \label{fig:teaser}
\end{figure}

High quality 3D human reconstruction is essential to large number of applications ranging from telecommunications, education, entertainment, and so on. High-end systems~\cite{url4dviews,url8i,bradley2008markerless,pons2017clothcap,dou2016fusion4d,Motion2Fusion} based on dense camera rigs (up to 100 cameras~\cite{collet2015high,joo2015Panoptic}) and custom-designed lighting conditions~\cite{VlasicPBDPRM09,guo2019relightables} can achieve high-quality reconstruction, but the sophisticated setup limits their deployment in practice. 
Recently, researchers have proposed to employ neural implicit functions as a learning approach to reconstruct 3D humans from single-view RGB~\cite{saito2019pifu,saito2020pifuhd}, sparse-view RGB~\cite{Huang_2018_ECCV,zheng2021deepmulticap} or sparse-view RGBD inputs~\cite{yu2021function4d}. Benifiting from the representation power of deep implicit functions, these works demonstrated visually pleasing results and inspired many follow-up works~\cite{zheng2021pamir,Huang2020ARCH,he2021arch++,li2021posefusion}.

Despite the significant progresses, reconstructing \emph{highly accurate} 3D human models from sparse-view ($<10$) passive RGB cameras is still far from a solved problem: the afore-mentioned learning-based approaches are still unable to reconstruct very high accurate results as even fed with dense inputs. 
The underlying reason is that these two types of methods take different cues to reconstruct 3D models. 
Current learning-based sparse-view methods mainly rely on the appearance cues and high-level semantics in each individual image to ``guess'' the geometry, neglecting the cross-view correspondence relationship. In contrast, dense-view systems explicitly perform stereo matching to establish dense correspondences across different views for analytical depth calculation, thus they can produce more accurate models without any data-driven prior. 
Therefore, we believe accurate stereo matching is the key to bring the sparse-view reconstruction quality to the next level, and we seek to provide a solution in this paper.

However, this is not an easy task considering the variations of 3D humans and the potential occlusions under sparse camera settings. 
Current state-of-the-art stereo methods, either for general purpose~\cite{lipson2021raft} or for human reconstruction~\cite{yang2021stereopifu}, typically assume close viewpoints and operate in a low resolution. Besides, their result quality is also limited by the discrete nature of pixel-wise cost volumes, failing to achieve sub-pixel accuracy. 
To address these challenges, we introduce a novel stereo formulation based on diffusion models~\cite{sohl2015deep,Ho2020Denoising}.
Diffusion models are a class of generative models designed for synthesizing data via modeling the gradient of data distribution. 
They can synthesize photo-realistic images and even beat GANs in terms of image quality~\cite{dhariwal2021diffusion}. 
The generative power of these models mainly owes to a natural fit to the inductive biases of image data when equipping a UNet as the nework backbone~\cite{rombach2021highresolution}. 
As mentioned in \cite{Song2021Score}, the diffusion process can be regarded as learning to solve a continuous stochastic differential equations, which, interestingly, coincides with traditional stereo methods based on a continuous variational formulation~\cite{liu2009Continuous}. 
Such similarity inspires us to propose \emph{diffusion-based human stereo}, a novel stereo method that combines the continuity of diffusion models with existing learning-based iterative stereo to achieve high-quality human depth estimation. 
To the best of our knowledge, our method is the first one that link the separate research threads of diffusion models and stereo in a synergistic architecture. 
As shown in the experiments, our diffusion-based stereo can produce high-quality human depth at the same accuracy level as traditional dense systems, while taking much less images as input.

Taking our diffusion-based human stereo as the core, we further present DiffuStereo, a novel system for high-quality human volumetric reconstruction from sparse views.
The key idea of our system is to utilize diffusion-based stereo to iteratively refine the geometry reconstructed by existing implicit representation-based methods, and this is done in a 2D flow domain. In particular, we first adopt DoubleField~\cite{shao2022doublefield} to reconstruct a coarse human mesh and render the coarse depth maps from multiple viewpoints. Then we transform them into disparity flows and compute the masks to identify possible occlusions. 
To deal with the challenge of stereo for sparse views and achieve sub-pixel accuracy for high-resolution inputs, we make two key designs as follows. Firstly, we condition the diffusion network with several features to ensure stereo consistency and add epipolar constraints to the predicted flow. Secondly, we design a two-level network structure to tackle with the memory issues for high-resolution images. The global level extracts human semantic features from the downsampled images, while the diffusion level introduces the global feature into our diffusion stereo network and iteratively refines the disparity flow. This two-level structure allows us to train the network in a patch-based manner and thus resolves the memory bottleneck caused by high-resolution inputs.
To fuse the the depth maps from different viewpoints into a 3D model, we propose a light-weight hybrid fusion strategy, which gracefully deals with calibration error on real-world data and completes the occluded region. This fusion step firstly aligns the refined depth point clouds and the coarse human model through non-rigid ICP. Then we select the points from the coarse mesh to complete the invisible regions and reconstruct the final model. 

Overall, our method requires only 8 passive cameras and can reconstruct high-quality human models at a level of detail that was never thought to be possible before. The quality of our geometry  reconstruction in visible regions is even competitive with the ground truth on THuman2.0 dataset~\cite{yu2021function4d}. 
% Moreover, our system runs in 15 seconds per frame, which is much faster than traditional methods based on dense camera rigs. 
In summary, our contributions in this work are:

\noindent 1) We propose DiffuStereo, a light-weight and high-quality system for human volumetric reconstruction under sparse multi-view cameras.

\noindent 2) To the best of our knowledge, we present the first method to introduce diffusion models into stereo and human reconstruction. We extend the vanilla diffusion model by carefully designing a new diffusion kernel and introducing additional stereo constraints into the diffusion conditions. 

\noindent 3) We propose a novel multi-level diffusion stereo network to achieve accurate and high-quality human depth estimation. Our network can gracefully handle high-resolution (up to 4k) images without suffering from memory overload. 

%% file: section_relwork.tex
\section{Related Work}
% \noindent 
\textbf{Stereo matching}
% Stereo matching 
aims at computing the disparity between two camera views. The classical pipelines typically consist of two stages, namely matching and filtering. In the literature, much attention has been paid to designing better matching cost~\cite{Hannah1974ComputerMO,Zabih1994Nonparametric} and better filter algorithms~\cite{Barnes2009PatchMatch,Hirschmuller2008Stereo,Kolmogorov2006Convergent}. 
With the advent of deep learning era, convolutional neural networks (CNNs) were introduced to improve pixel matching in the stereo pipeline~\cite{zbontar2015computing,Mayer2016dataset}.  
To fully realize the potential of deep networks, researchers proposed many end-to-end stereo architectures~\cite{Chang2018Pyramid,guo2019group,Hirschmuller2008Stereo,Zhang2019GANet,Kendall2017Endtoend,Yao2018MVSNet,Zhang2020Domain,Zhang2020Adaptive}. 
At the core is a 3D cost volume constructed over 2D feature maps, followed by 3D convolutional layers for correspondence filtering. 
However, such architectures come at a high computational cost and limits the possible operating resolution. Recently, RAFT-stereo~\cite{lipson2021raft} proposed a memory-efficient stereo method by replacing 3D convolutions with 2D ones and predicting disparity in an iterative manner. Other methods~\cite{li2022practical,wang2022itermvs,wang2022efficient} formulating stereo matching in an iterative process further improve the geometry accuracy and generalization ability. Although these methods achieve impressive results on datasets such as KITTI~\cite{Geiger2012CVPR,Menze2015CVPR} and DTU~\cite{jensen2014large},
% and FlyingThings3D~\cite{Mayer2016dataset}
we fount that they cannot work well in our sparse-view, high-resolution setting. Therefore, we design a novel diffusion-based stereo method for sparse-view human reconstruction.

% \noindent 
\textbf{Traditional 3D human reconstruction}
% Reconstructing human-centered scenes from images 
% has been an active research area in computer vision. 
% Previous studies on reconstructing human-centered scenes  focused on using 
methods rely on
multi-view images~\cite{StarckCGA07,liu2009point,WuShadingHuman,VlasicPBDPRM09} or RGB(D) image sequences~\cite{alldieck2018videoavatar,alldieck2018videoavatar_detailed,yu2017BodyFusion,yu2018doublefusion,Zheng2018HybridFusion,bogo2015detailed,dou2016fusion4d,Motion2Fusion,yu2021function4d} to reconstruct the geometric model. Extremely high-quality reconstruction results have also been demonstrated with a large amount of cameras~\cite{collet2015high,guo2019relightables}. 
The most essential part in these pipelines is the depth point cloud obtained using classical stereo matching, and researchers have adopted various technologies to further improve the accuracy of depth estimation, such as photometric stereo under different illuminations~\cite{VlasicPBDPRM09} or deep learning on active stereo patterns~\cite{fanello2017ultrastereo}. 
Variational formulation was also proposed for the purpose of continuous depth computation and excels at detail capture~\cite{liu2009Continuous}, but solving such a variational energy easily falls into local minimals, resulting into poor robustness. 
We overcome this problem by employing diffusion models, which solves the problem of variational depth estimation with data-driven knowledge and demonstrates robust performance.

\textbf{Learning-based 3D human reconstruction}
are recently proposed in order to reduce the difficulty in system setup. 
They learn a data-driven prior from high-quality 3D human database and reconstruct 3D humans from sparse camera views~\cite{MinimalCam18,Huang2018SparseView,shao2022doublefield} or even single-view images~\cite{natsume2019siclope,saito2019pifu,saito2020pifuhd,zheng2021pamir,gabeur2019moulding,wang2020normalgan,zhu2019hmd,alldieck2019tex2shape,Huang2020ARCH,he2021arch++,li2021posefusion}. 
For example, PIFu~\cite{saito2019pifu} and PIFuHD~\cite{saito2020pifuhd} proposed to regress a deep implicit function using pixel-aligned image features and is able to reconstruct high-resolution results. 
% ARCH~\cite{Huang2020ARCH} and ARCH++~\cite{he2021arch++} proposed to reconstruct 3D human models in a canonical pose in order to support animation. StereoPIFu~\cite{yang2021stereopifu} showed that combining stereo and implicit reconstruction can further improve the accuracy of geometry. 
% POSEFusion~\cite{li2021posefusion} introduced a keyframe selection scheme to recover from monocular input the dynamic details in invisible regions. 
DoubleField~\cite{yu2018doublefusion} combined the merits of implicit geometric representations and radiance fields for high-fidelity human reconstruction and rendering.
In spite of the progress in these works, challenges still remain in reconstructing highly accurate human models from sparse views, and we identify accurate, high-resolution stereo matching for sparse views as the key to this problem.

% Although demonstrating plausible results, these methods rely on large scale dataset of 3D human scans to train the model, and suffer from reconstruction errors and weak generalization capability. In contrast, our method bypasses the reconstruction step and directly learns an animatable avatar from RGB videos. 

% \noindent 
\textbf{Diffusion Models}
are a class of generative models based on a Markov chain~\cite{sohl2015deep,Ho2020Denoising,Song2021Denoising}. They convert samples from a standard Gaussian distribution into ones from an empirical data distribution through an iterative denoising process. 
The denoising process can be extended for the purpose of conditional generation by adding other signals as the condition~\cite{Chen2021WaveGrad}. 
When implementing the network backbone as a UNet, diffusion models are well suited for image-like data and achieved state-of-the-art results in image generation~\cite{Nichol2021Improved,rombach2021highresolution,dhariwal2021diffusion}, super-resolution~\cite{saharia2021image,ho2021cascaded,Li2022srdiff}, image-to-image translation~\cite{Saharia2021Palette} and so on. 
In this paper, we show that diffusion models, with some slight modifications, can also be used in sparse-view stereo matching for human reconstruction.

% Very recently, diffusion model has been proposed to synthesis high-quality and photo-realistic images~\cite{}. The powerful generative ability of diffusion model even beats GAN~\cite{}. The key insight of diffusion model is to learn the gradient of data distribution. As mentioned in ~\cite{}, the diffusion process can be regarded as learning to solve a continuous stochastic differential equations. Meanwhile, traditional stereo methods based on continuous variational approach~\cite{} also solve the differential equations to overcome the discretization limitations and estimate the sub-pixel level flow. Such similarity inspires us to combine the continuity of diffusion model with existing learning-based discrete stereo to achieve high quality human depth estimation. 

%% file: section_method.tex
\section{Method}
\begin{figure}[t]
    \centering
    \includegraphics[width=\linewidth]{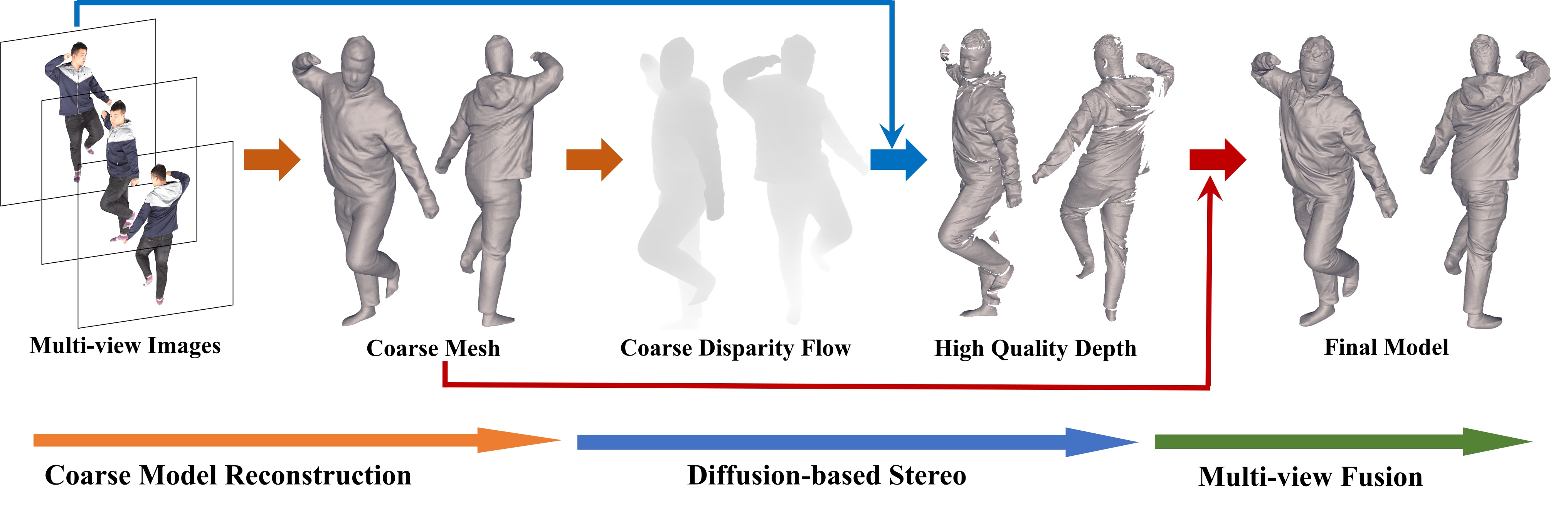}
    \caption{\textbf{Overview of the DiffuStereo system.} Our system consists of three key steps to reconstruct high-quality human models from sparse-view inputs: i) An initial human mesh is predicted by DoubleField~\cite{shao2022doublefield}, and rendered as the coarse disparity flow (Sec.~\ref{sec:initial}); ii) The coarse disparity maps are refined in the diffusion-based stereo to obtain the high-quality depth maps (Sec.~\ref{sec:diff_stereo}); iii) The initial human mesh and high-quality depth maps are fused as the final high-quality human mesh (Sec.~\ref{sec:fusion}).}
    \label{fig:pipeline}
\end{figure}

As shown in Fig.~\ref{fig:pipeline}, our system DiffuStereo can reconstruct high-quality human models from sparse (as few as 8) cameras.
All cameras evenly distribute on a ring surrounding the target human.

Such a sparse setting poses huge challenges to high-quality reconstruction as the angle between two neighboring views can be as large as $45^{\circ}$.
DiffuStereo tackles the challenges with the joint efforts of an off-the-shelf reconstruction method DoubleField~\cite{shao2022doublefield}, a diffusion-based stereo network, and a light-weight multiview fusion module.
Overall, DiffuStereo consists of three key steps to reconstruct high-quality human models from sparse-view inputs: 

i) An initial human mesh is predicted by DoubleField~\cite{shao2022doublefield}, and rendered as the coarse disparity maps (Sec.~\ref{sec:initial}); In DoubleField, one of the most recent state-of-the-art human reconstruction methods, the surface and radiance fields are bridged to leverage human geometry priors and multi-view appearances, providing a good initialization of mesh given sparse-view inputs.

ii) For each two neighboring views, the coarse disparity maps are refined in the diffusion-based stereo network to obtain the high-quality depth maps (Sec.~\ref{sec:diff_stereo}); The diffusion-based stereo network has a strong capability to improve the disparity maps for each input views, where a diffusion process is employed for continuous disparity refinements.

iii) The initial human mesh and high-quality depth maps are fused into the final high-quality human mesh (Sec.~\ref{sec:fusion}), where a light-weight multiview fusion module takes the initial mesh as the anchor position and effectively assembles the partial refined depth maps. 

\subsection{Mesh, Depth, Disparity Initialization}
\label{sec:initial}

Given a set of $N$-view images $\{\mathbf{I}^1,\dots,\mathbf{I}^N\}$, an initial human mesh $\mathbf{m}_c$ is predicted by DoubleField~\cite{shao2022doublefield} and rendered as the coarse depth maps $\{\mathbf{D}_c^1,\dots,\mathbf{D}_c^N\}$ for $N$ input viewpoints. These depth maps are further transformed into the disparity maps as they are necessary for stereo matching. Without loss of generality, let $m$ and $n$ be the index of two neighboring views. To obtain the coarse disparity map $\mathbf{x}_c$ from the view $m$ to its neighboring view $n$, we take the depth map $\mathbf{D}_c^{m}$ of the view $m$ and compute disparity at the pixel position $\bm{o}=(i,j)$ as following:
\begin{align}
\label{eqn:flow_projection}
    \mathbf{x}_c(\bm{o}) = \mathbf{\pi}^{n}\left(\left(\mathbf{\pi}^{m}\right)^{-1}\left(\left[\bm{o}, \mathbf{D}_c^{m}(\bm{o})\right]^\mathrm{T}\right)\right) - \bm{o}
\end{align}
where $(\mathbf{\pi}^{m})^{-1}$ transforms the points from the depth map $\mathbf{D}_c^{m}$ to the world coordinate system and $\mathbf{\pi}^{n}$ projects the points in the world coordinate system to the image coordinate system.

Since the initial disparity maps are calculated from a coarse human mesh, the large displacement and occlusion region issues can be largely alleviated. As will be presented shortly, these disparity maps are further refined by a Diffusion-based Stereo to obtain high-quality depth maps for each input viewpoint.

\subsection{Diffusion-based Stereo for Disparity Refinement}
\label{sec:diff_stereo}
Existing stereo methods~\cite{Chang2018Pyramid,guo2019group,Hirschmuller2008Stereo,Zhang2019GANet,Kendall2017Endtoend,Yao2018MVSNet,Zhang2020Domain,Zhang2020Adaptive} adopt 3D/4D cost volumes to predict disparity map in a discrete manner, which is difficult to achieve sub-pixel level flow estimation. 
To overcome this limitation, we propose a diffusion-based stereo such that the stereo network can learn a continuous flow during the iterative process.
Specifically, our diffusion-based stereo contains a forward process and a reverse process to obtain the final high-quality disparity map.
In the forward process, the initial disparity maps are diffused to the maps with noise distribution.
In the reverse process, the high-quality disparity maps will be recovered from the noisy maps with the condition of several stereo-related features.
In the following, we give a brief introduction of generic diffusion models and then introduce our solution to combine the continuity of diffusion models and the learning-based iterative stereo. 
Moreover, we also present a multi-level network structure to resolve the memory issues for high-resolution images input.

\subsubsection{3.2.1 Generic Diffusion Model}

The standard T-step diffusion model~\cite{Ho2020Denoising} contains a forward process and a reverse process. The forward process is about gradually adding Gaussian noise to the original input $\mathbf{y}_0$ in each step $t$ such that they are turned into the pure noise $\mathbf{y}_T$ at step $T$. The reverse process is about recovering $\mathbf{y}_0$ from the noise $\mathbf{y}_T$ iteratively, which can be regarded as denoising.
More formally, the diffusion model can be written as two Markov Chains:
\begin{align}
    \label{eqn:diff_forward}
    q(\mathbf{y}_{1:T}|\mathbf{y}_0) & = \prod_{t=1}^T q(\mathbf{y}_t|\mathbf{y}_{t-1}),\\
    \label{eqn:diff_kernel}
    q(\mathbf{y}_t|\mathbf{y}_{t-1}) & = \mathcal{N}(\sqrt{1-\beta_t}\mathbf{y}_{t-1}, \beta_t I)\\
    \label{eqn:diff_reverse}
    p_\theta(\mathbf{y}_{0:T}|\mathbf{s}) & = p(\mathbf{y}_T)\prod_{t=1}^Tp_\theta(\mathbf{y}_{t-1}|\mathbf{y}_t, \mathbf{s}),
\end{align}    
where $q(\mathbf{y}_{1:T}|\mathbf{y}_0)$ is the forward function, $q(\mathbf{y}_t|\mathbf{y}_{t-1})$ is the diffusion kernel representing the way to add noise, $\mathcal{N}$ the normal distribution, $\mathbf{I}$ the identical matrix, $p_\theta()$ the reverse function which adopts a denoising network $\mathcal{F}_{\theta}$ to denoise $\mathbf{y}_{T}$ and $\mathbf{s}$ the additional condition.
When $T\rightarrow \infty$, the forward and the reverse process can be seen as a continuous process or stochastic differential equations~\cite{Song2021Score}, which is a natural form for continuous flow estimation.
As validated in previous work~\cite{Song2021Score}, injecting Gaussian noise into the parameter updates makes the iterative process more continuous and can avoid collapses into local minimal.
In this work, we will show that, such a powerful generative tool can be also leveraged to produce continuous flows for the human-centric stereo task.

Compared with the generic diffusion models, two task-specific designs are adopted in our diffusion-based stereo: i) a new diffusion kernel is used in consideration that the stereo flow estimation is not purely a generative task; ii) stereo-related features and supervisions are involved in the reverse process to ensure the color consistency and epipolar constraints.

\subsubsection{3.2.2 Disparity Forward Process}
\label{sec:diffusion_process}

In diffusion-based stereo, the forward process gradually transforms the distribution of disparity flows to the noisy distribution.
Specifically, the input $\mathbf{y}_0$ of the diffusion model in our case is the residual disparity $\mathbf{\hat{y}}_0$ between the ground truth disparity maps $\hat{\mathbf{x}}$ and the coarse disparity maps $\mathbf{x}_c$, \ie, $\mathbf{\hat{y}}_0=\hat{\mathbf{x}}-\mathbf{x}_c$.
Different from existing generative diffusion model for image synthesis which utilizes $\sqrt{1-\beta_t}$ to gradually reduce the scale of $\mathbf{y}_{t-1}$, we 
design a diffusion kernel to preserve the scale of $\mathbf{y}_{t-1}$ and linearly drifts $\mathbf{y}_0$ to $\mathbf{y}_t$, \ie, Eqn.~\eqref{eqn:diff_kernel} is rewritten as:
\begin{align}
\label{eqn:new_kernel}
   q(\mathbf{y}_t|\mathbf{y}_{t-1})& =\mathcal{N}(\mathbf{y}_t|\mathbf{y}_{t-1} - \alpha_t \mathbf{y}_0, \alpha_t\mathbf{I}),
\end{align}    
where $\alpha_t$ is the parameter to scale the noise.
Based on this new diffusion kernel, the disparity forward process gradually adds noise to the ground truth residual disparity $\hat{\mathbf{y}}_0$ using Eqn.~\eqref{eqn:diff_forward} and~\eqref{eqn:new_kernel}.
In our experiments, we find our diffusion kernel makes the reverse process more stable and efficiently reduces the needed number of iterative steps at inference. Derivation of the forward process under our new kernel is similar with~\cite{Ho2020Denoising} and can be found in Supp.Mat.

\begin{figure}[t]
    \centering
    \includegraphics[width=.95\linewidth]{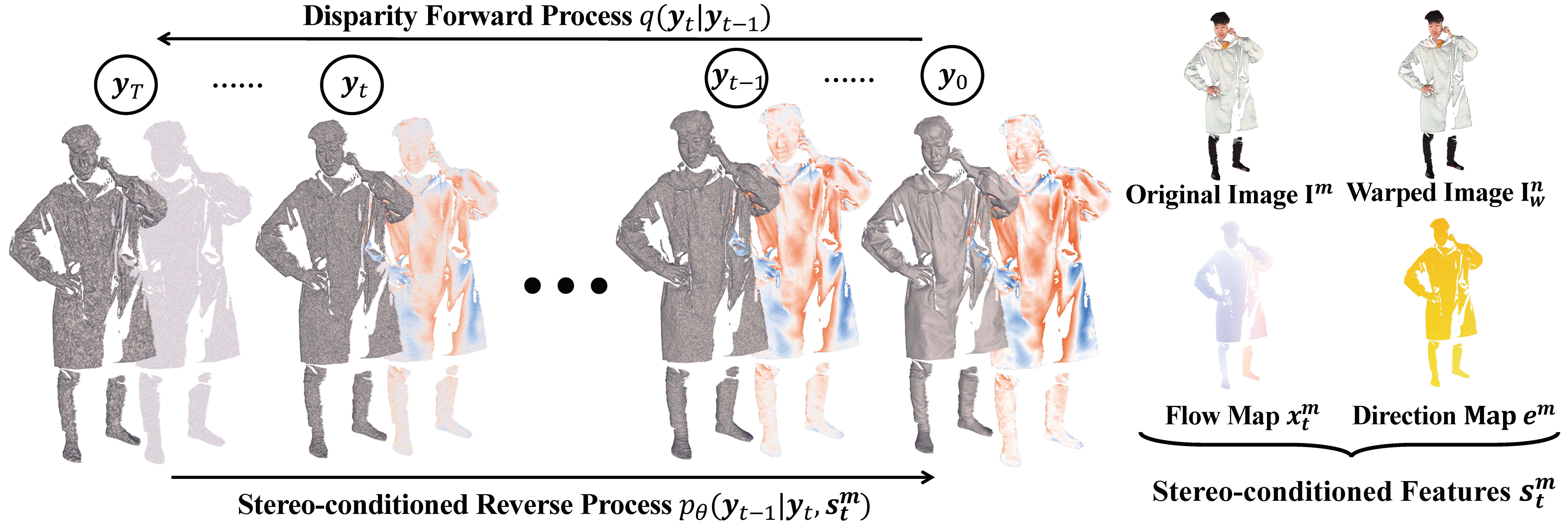}
    \caption{\textbf{Illustration of the forward process and the reverse process in our diffusion-based stereo.} In the forward process, the initial residual disparity maps $\mathbf{y}_{T}$ are diffused to the noisy maps $\mathbf{y}_0$. In the reverse process, the high-quality residual disparity maps will be recovered from the noisy maps with the condition of several stereo-related features $\mathbf{s}^{m}_t$.}
    \label{fig:diffusion}
\end{figure}

\subsubsection{3.2.3 Stereo-conditioned Reverse Process.}
\label{sec:conditional_diffusion}

The reverse process of diffusion-based stereo aims at recovering the residual disparity $\hat{\mathbf{y}}_0$ from the noise $\mathbf{y}_T$ using Eqn.~\eqref{eqn:diff_reverse}.
In this process, a diffusion stereo network acts as the denoising network $\mathcal{F}_\theta$ in Eqn.~\eqref{eqn:diff_reverse} by taking $\mathbf{y}_t$ and $\mathbf{s}$ as inputs and predict $\widetilde{\mathbf{y}}_0$.

As the diffusion kernel also influences the reverse process, the denoising process of our network is different from the generic one under our new kernel.
Given the kernel in Eqn.~\eqref{eqn:new_kernel}, the formulation of each step in the reverse process can be written as:
\begin{align}
p_\theta(\mathbf{y}_{t-1}|\mathbf{y}_{t}, \mathbf{s}) = \mathcal{N}(\mathbf{y}_{t-1}|\mu_\theta(\mathbf{y}_t, \gamma_t, \mathbf{s}), \sigma_t^2\mathbf{I}),
\end{align}
where $\gamma_t=\sum_{i=1}^t \alpha_i$, $\sigma_t^2=\frac{\alpha_t\gamma_{t-1}}{\gamma_t}$, and $\mu_\theta()$ is the prediction process of the denoising network $\mathcal{F}_\theta$.
Moreover, we substitute the predicted $\widetilde{\mathbf{y}}_0$ into the posterior distribution of $q(\mathbf{y}_{t-1}|\mathbf{y}_t, \mathbf{y}_0)$ to represent the mean of $p_\theta(\mathbf{y}_{t-1}|\mathbf{y}_t)$:
\begin{align}
    \mu_\theta(\mathbf{y}_t, \gamma_t, \mathbf{s})=\frac{\alpha_t}{\gamma_t}\widetilde{\mathbf{y}}_0+\frac{\gamma_{t-1}}{\gamma_t}\mathbf{y}_t.
\end{align}
In this way, the whole reverse process can be formulated as:
\begin{align}
\label{eqn:reverse}
\mathbf{y}_{t-1} \leftarrow \frac{\alpha_t}{\gamma_t}\widetilde{\mathbf{y}}_0+\frac{\gamma_{t-1}}{\gamma_t}\mathbf{y}_t +  \frac{\alpha_t\gamma_{t-1}}{\gamma_t}\epsilon_t, \epsilon_t \sim \mathcal{N}(\mathbf{0}, \mathbf{1}).
\end{align}

Since the disparity refinement is not a fully generative process, the diffusion stereo network $\mathcal{F}_\theta$ takes additional conditions as inputs to recover the high quality disparity flows.
In our solution, four types of stereo-related maps (Fig.~\ref{fig:diffusion} Right) act as the additional conditions $\mathbf{s}$ in Eqn.~\eqref{eqn:diff_reverse} to ensure the color consistency and epipolar constraints at each step of the reverse process:

1) The original image $\mathbf{I}^{m}$ of the view $m$;

2) The warped image $\mathbf{I}_w^{n}$ of view $n$, which is obtained by transforming pixels of $\mathbf{I}^{n}$ using current flow $ \mathbf{x}_t^{m} =  \mathbf{x}^{m} + \mathbf{y}_t$:
\begin{align}
    \mathbf{I}_w^{n}(\bm{o}) = \mathbf{I}^{n}(\mathbf{x}_t^{m}(\bm{o}) + \bm{o}).
\end{align}

3) The current flow map $\mathbf{x}_t^{m}$;

4) The direction map $\mathbf{e}^{m}$ of epipolar line, which is computed as:
\begin{equation}
\mathbf{e}^{m} = (\dot{\mathbf{x}}^{m}_c - \mathbf{x}_c^{m}) / \| \dot{\mathbf{x}}^{m}_c - \mathbf{x}_c^{m}\|_2,
\end{equation}
where $\dot{\mathbf{x}}^{m}_c$ is the shifted flow map transformed based on the coarse depth map $\mathbf{D}^{m}_c$ and a fixed shift $\beta$:
\begin{equation}
\dot{\mathbf{x}}^{m}_c(\bm{o}) = \mathbf{\pi}^{n}((\mathbf{\pi}^{m})^{-1}(\left[\bm{o}, \mathbf{D}_c^{m}(\bm{o})+\beta \right]^T)) - \bm{o}.
\end{equation}

Among the above four conditions, $\mathbf{I}^{m}$ and $\mathbf{I}_w^{n}$ encourage the network to be aware of color consistency, while $\mathbf{x}_t^{m}$ and $\mathbf{e}^{m}$ provide the hints about the flow movement to network for better predictions.
We condition the network by concatenating the above stereo-related maps as $\mathbf{s}^{m}_t=\bigoplus(\mathbf{I}^{m}, \mathbf{I}_w^{n}, \mathbf{x}^{m}_t, \mathbf{e}^{m})$.
The conditions $\mathbf{s}^{m}_t$ are further concatenated with $(\mathbf{y}_t^{m}, t)$ and fed into the diffusion stereo network. 
Additionally, we also constrain the network output map $\bm{R}$ to only one channel such that the predicted residual flow $\mathbf{\widetilde{y}}_t=\mathbf{e}^{m}\cdot \bm{R}$ is forced to move along the epipolar line.

When the reverse process is finished, the final flow $\mathbf{x}^{m}+\widetilde{\mathbf{y}}_0^{m}$ will be transformed back to the refined depth map $\mathbf{D}_f^{m}$ of the view $m$ using the inverse formulation of Eqn.~\eqref{eqn:flow_projection}.

\begin{figure}[t]
    \centering
    \includegraphics[width=\linewidth]{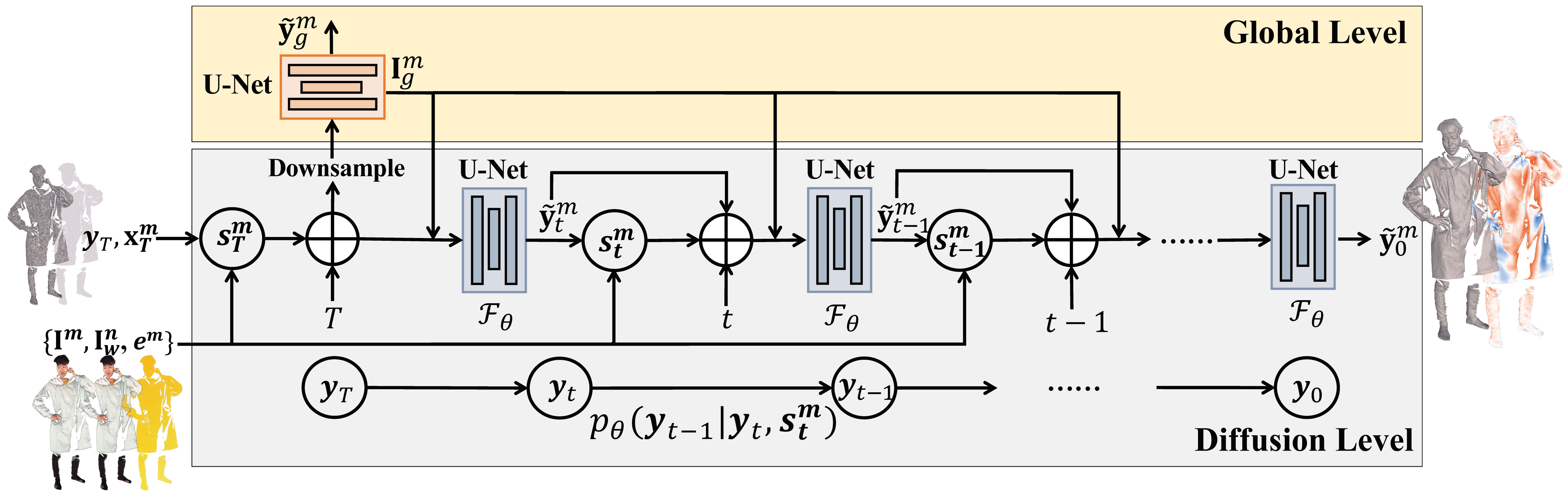}
    \caption{\textbf{The proposed multi-level stereo network structure.} 
    A global network $\mathcal{F}_g$ is combined with the diffusion stereo network $\mathcal{F}_\theta$ so that our method can leverage information at both global and diffusion levels to predict high-quality depths. 
 }
    \label{fig:network}
\end{figure}

\subsubsection{3.2.4 Multi-level Network Structure}
\label{sec:two_level}

For high quality human reconstruction, it is essential to leverage high-resolution input images.
When applying the above diffusion stereo network, the memory issues arise when high-resolution images (4K in our experiments) are taken as inputs.
Inspired by PIFuHD~\cite{saito2020pifuhd}, we tackle this issue using a multi-level network structure, in which a global network $\mathcal{F}_g$ is combined with the diffusion stereo network $\mathcal{F}_\theta$.
In this way, $\mathcal{F}_g$ and $\mathcal{F}_\theta$ can produce the disparity flow at the global level and the diffusion level, respectively. 

As shown in Fig.~\ref{fig:network}, at the global level, both the coarse initial flow $\mathbf{x}_c^{m}$ and the conditional stereo features $\bm{s}^{m}$ are downsampled to the resolution of $512\times512$ and then fed into the global network $\mathcal{F}_g$.
which directly predicts a low-resolution residual flow $\widetilde{\mathbf{y}}_g^{m}$. 
Note that the flow estimation at the global level is not a diffusion process but it learns global features which contains human semantic information. 
At the diffusion level, we adopt the last image features $\mathbf{I}_g^{m}$ from the global network as an additional condition for the diffusion stereo network. 
Thus, the stereo-related features in the local-level network is modified to $\mathbf{s}^{m}_t=\bigoplus(\mathbf{I}^{m}, \mathbf{I}_w^{n}, \mathbf{x}^{m}_t, \mathbf{e}^{m}, \mathbf{I}_g^{m})$. 
Benefiting from the multi-level structure, the memory issues can be largely overcomed as the diffusion stereo network can be trained in a patch-based manner. Moreover, with the guidance of the global feature $\mathbf{I}_g^{m}$, our diffusion stereo network can focus more on the recovery of fine details.

\subsubsection{3.2.5 Training of Diffusion-based Stereo}
\label{sec:training}

\noindent \textit{Global Level.} We downsample the ground truth residual flow to the resolution of $512$ and train the global network $\mathcal{F}_g$ with the global loss $\mathcal{L}_g$:
\begin{align}
\mathcal{L}_g = \frac{1}{HW}\sum_{i=1}^{H}\sum_{j=1}^{W}\|\widetilde{\mathbf{y}}_g(i, j) - \mathbf{y}_0(i, j) \|^2,
\end{align}
The global loss encourages the network to learn human semantic features for the flow estimation.

\noindent \textit{Diffusion Level.} Following ~\cite{Chen2021WaveGrad}, we generate training samples by randomly selecting a time step $t$ and diffusing the ground truth residual flow $\mathbf{y}_0$ to $\mathbf{y}_t$ using:
\begin{align}
\label{eqn:sampling}
    q(\mathbf{y}_t|\mathbf{y}_0) = \mathcal{N}(\mathbf{y}_t|(1-\gamma_t)\mathbf{y}_{0}, \gamma_t\mathbf{I}).
\end{align}
Then we adopt diffusion loss $\mathcal{L}_d$ to train our diffusion stereo network $\mathcal{F}_\theta$ at the $t$-th step:
\begin{align}
\mathcal{L}_d = \frac{1}{HW}\sum_{i=1}^{H}\sum_{j=1}^{W}\|(\mathcal{F}_\theta(\mathbf{y}_t, \mathbf{s}_t, t))(i, j) - \mathbf{y}_0(i, j) \|^2_2.
\end{align}

\subsection{Light-weight Multi-view Fusion}
\label{sec:fusion}
In this section, we propose a light-weight multi-view hybrid fusion to fuse the refined depth maps $\mathbf{D}_f^{1}, ..., \mathbf{D}_f^{n}$ and the coarse human mesh $\mathbf{m}_c$ to reconstruct the final model.
Before fusion, we first remove the depth boundaries using an erosion kernel and transform each refined depth map $\mathbf{D}^{i}_f$ to a point cloud $\mathbf{p}^{i} = (\mathbf{\pi}^{i})^{-1}(\mathbf{D}^i_f)$.
Since calibration error is inevitable in real-world data, the refined depth maps estimated from multi-view may not be accurately aligned. 
To handle this issue, we utilize non-rigid ICP to align the depth point cloud $\mathbf{p}^{i}$ and the point cloud $\mathbf{p}^c$ of the coarse mesh, where the coarse point cloud serves as the anchor model for subsequent alignment. 
In our non-rigid ICP, the optimization objectives $\mathcal{L}_{icp}=\mathcal{L}_d + \mathcal{L}_s$ consist of the data term $\mathcal{L}_d$ and the smooth term $\mathcal{L}_s$, which are defined as following:
\begin{align}
      \mathcal{L}_d & = \sum_{i=1}^n \sum_{j=i+1}^n \sum_{(a, b) \in \widetilde{\mathbf{N}}^{i, j}}\|\widetilde{\mathbf{p}}^{i}_{ a}-\widetilde{\mathbf{p}}^{j}_{b}\|^2 + \lambda_d \sum_{i=1}^n\sum_{(a, b) \in \widetilde{\mathbf{N}}^{i}_{c}} \|\widetilde{\mathbf{p}}^{i}_{a}-\mathbf{p}^c_{b} \|^2 \\
      \mathcal{L}_s & = \lambda_s \sum_{i=1}^n \sum_{(a, b) \in \mathbf{N_i}}\|\widetilde{\mathbf{d}}^{i}_{a} - \widetilde{\mathbf{d}}^{i}_{b} \|^2 / \|\mathbf{p}^{i}_{a}-\mathbf{p}^{i}_{b} \|^2,
\end{align}
where $\mathbf{d}^{i}$ is the displacement of the depth point cloud $\mathbf{p}^{i}$, $\mathbf{N}$ is the set of the searched neighborhood correspondence. We adopt the nearest neighbour algorithm to search correspondence and the threshold of search radius is 2mm.

After optimization, the final point cloud $\mathbf{p}^{f}$ is the union of optimized depth point cloud $\widetilde{\mathbf{p}}^{i}$ and the coarse point cloud $\mathbf{p}^c$, while the final mesh can be reconstructed from the final point cloud $\mathbf{p}^f$ using Poisson Reconstruction~\cite{kazhdan2006poisson}.

%% file: section_exps.tex
\section{Experiment}

\subsection{Implementation Details}
In our implementation, we adopt a U-Net~\cite{ronneberger2015unet} model which is similar with~\cite{dhariwal2021diffusion} as the structure of the global network $\mathcal{F}_g$ and the diffusion network $\mathcal{F}_\theta$.
The $T$ in our diffusion model is $30$ which is much less than the original model in image generation tasks.
For the other diffusion parameters including $\alpha_t, \gamma_t$ and more implementation details, please refer to the Supplementary Material.

\noindent \textbf{Training Data Preparation.}
We collect 300 models from Twindom~\cite{twindom} and render images pairs for training. 
We first render images and depth maps with 4K resolution densely from 360° angles.
We then run DoubleField~\cite{shao2022doublefield} on the images of 8 even-distributed views to predict the SDF volume with a resolution of $512^3$ and further retrieve the coarse human mesh using marching cube.
During the training of our diffusion-based stereo network, we randomly select two views from the rendered images of the same model and constrain their
angle in the interval of $[20,50]$.
We also compute occlusion regions between two views and filter out bad parts to avoid unstable training. 

\noindent \textbf{Evaluation Data Preparation.}
We randomly pick 300 models from the THUman2.0~\cite{yu2021function4d} dataset for evaluation. Person images and depth maps with 4K resolution are rendered from 360 angles.

\begin{table}[t]
   \scriptsize
   \centering
   \caption{Qualitative comparisons of stereo matching on the THuman2.0 dataset.}
   \label{tab:stereo_table}
   \begin{tabular}{p{3.5cm}|ccc|ccc|ccc|ccc}
  \hline
  \multirow{2}*{Method} 
  & \multicolumn{3}{c|}{\makecell[c]{AvgErr}}
  & \multicolumn{3}{c|}{\makecell[c]{1/2pix (\%)}} 
  & \multicolumn{3}{c|}{\makecell[c]{1pix (\%)}}
  & \multicolumn{3}{c}{\makecell[c]{3pix (\%)}} \\
         & 20° & 30° & 45° & 20° & 30° & 45° & 20° & 30° & 45° & 20° & 30° & 45° \\ \hline
        Stereo-PIFu~\cite{yang2021stereopifu} & 12.03 & 14.57 & 17.92 & - & - & - & - & - & - & - & - & - \\
       Raft-Stereo~\cite{lipson2021raft}   & 11.79 & 13.95 & 15.42 & - & - & - & - & - & - & - & - & - \\
      
     Stereo-PIFu(w. $\mathbf{D}_c$)  & 1.052 & 1.411 & 2.248 & 45.6  & 32.8 & 18.6 & 68.1  & 56.1 & 35.4 & 92.2 & 87.9 & 77.2 \\
       Raft-Stereo (w. $\mathbf{D}_c$)  & 0.812 & 1.020 & 1.567 & 54.0 & 44.1 & 26.1 & 74.9 & 68.6 & 48.0 & 95.1 & 93.4 & 89.7 \\ 
      Our method(w./o. diff.) & 0.799 & 0.933 & 1.448 & 56.2 & 47.2 & 28.2 & 76.9 & 71.6 & 51.1 & 95.4 & 94.4 & 89.4 \\ 
      Our method(Origin. diff.) & 0.618 & 0.704 & 0.881 & 62.5 & 57.8 & 52.9 & 85.2 & 81.5 & 77.2 & 97.5 & 96.7 & 94.8 \\  
       Our method & \textbf{0.483} & \textbf{0.515} & \textbf{0.632} & \textbf{71.3} & \textbf{70.0} & \textbf{67.6} & \textbf{90.6} & \textbf{89.0} & \textbf{85.9} & \textbf{98.6} & \textbf{97.9} & \textbf{96.6} \\  \hline
  \end{tabular}
\end{table}

\subsection{Comparisons on Stereo}
\begin{figure}[t]
    \centering
    \includegraphics[width=.9\linewidth]{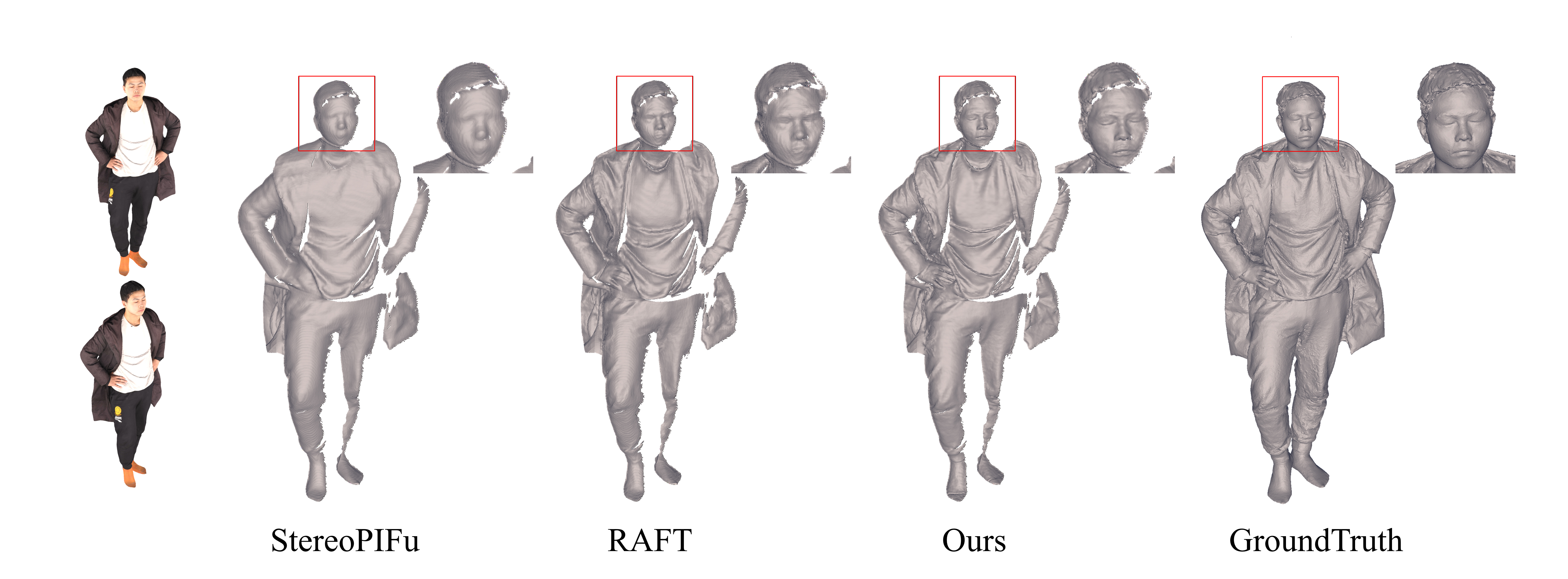}
    \caption{\textbf{Comparison in terms of depth estimation quality.} Compared to state-of-the-art method, ours can produce more geometric details, which is important for subsequent mesh reconstruction.}
    \label{fig:comparison_stereo}
\end{figure}

We quantitatively and qualitatively compare our diffusion-based stereo network with the state-of-the-art stereo method RAFT-Stereo~\cite{lipson2021raft} and the stereo matching network in StereoPIFu~\cite{yang2021stereopifu} based on 3D convolution. For each synthetic human model, we select 4 pairs of views with the view angle randomly chosen from (20°, 30°, 45°) and evaluate the performance of stereo. We also prepare the initial disparity flow based on the coarse mesh predicted by DoubleField~\cite{shao2022doublefield} for RAFT-Stereo and StereoPIFu. For quantitative comparisons, we follow RAFT-Stereo~\cite{lipson2021raft} to adopt the same metrics, \ie, the average error in pixels and the ratio of pixel error in three levels (1/2 pix, 1 pix and 3 pix). As shown in Tab.~\ref{tab:stereo_table}, our method significantly beats the other two methods in all metrics and achieves the best performance especially at the sub-pixel level. There are 71.3, 70.0, and 67.6 percent of our disparity estimation results within 1/2 pixel error for view angles of 20°, 30°, and 45°, respectively. More importantly, when two views become sparser, the sub-pixel level performance of our method remains very strong (only a decrease of 3.7$\%$) compared with the other two methods (decrease of 27$\%$ and 27.9$\%$), which validates the efficiency of our diffusion stereo network and the powerful ability of diffusion model for continuous disparity flow estimation. We also compare our method with the original RAFT-Stereo and StereoPIFu without initial value $\mathbf{D}_c$. However, they all fail to estimate disparity flow with sparse-view inputs and result in large average errors ($\geq$10 pixel). Hence, we only report the average error in pixels for StereoPIFu and RAFT-Stereo under this setting.
We also report qualitative results on THuman2.0 dataset in Fig.~\ref{fig:comparison_stereo} and the Supp.Doc. Compared with StereoPIFu and RAFT-Stereo, our method recovers more geometry details and the quality of depth is even competitive to the ground truth. Such a high-quality depth estimation is also benefited from our diffusion-based stereo and the design of multi-level network structure, as demonstrated in our ablation study.

\begin{table}[t]
   \scriptsize
   \centering
   \caption{Quantitative human geometry reconstruction results.}
   \label{tab:recon_table}
   \begin{tabular}{p{4cm}|ccccc}
  \hline
  \multirow{2}*{Method} 
  & \multicolumn{5}{c}{\makecell[c]{THuman2.0 \small{(8 views)}}} \\
         & Chamfer & P2S & 1mm(\%) & 2mm(\%) & 5mm(\%) \\ \hline
         PIFuHD~\cite{saito2020pifuhd}   & 3.018 & 2.509 & 21.0 & 58.9  & 90.0 \\
      StereoPIFu~\cite{yang2021stereopifu}   & 2.629 & 2.251 & 26.0 & 64.7 & 91.7 \\
        DoubleField~\cite{shao2022doublefield} & 2.879 & 2.389 & 23.2  & 61.6 & 90.8 \\
        Our Method &  \textbf{1.198} & \textbf{1.258} & \textbf{68.1} & \textbf{91.9} & \textbf{96.6} \\   \hline
  \end{tabular}
\end{table}

\subsection{Comparisons on Human Reconstruction}
\begin{figure}[t]
    \centering
    \includegraphics[width=.9\linewidth]{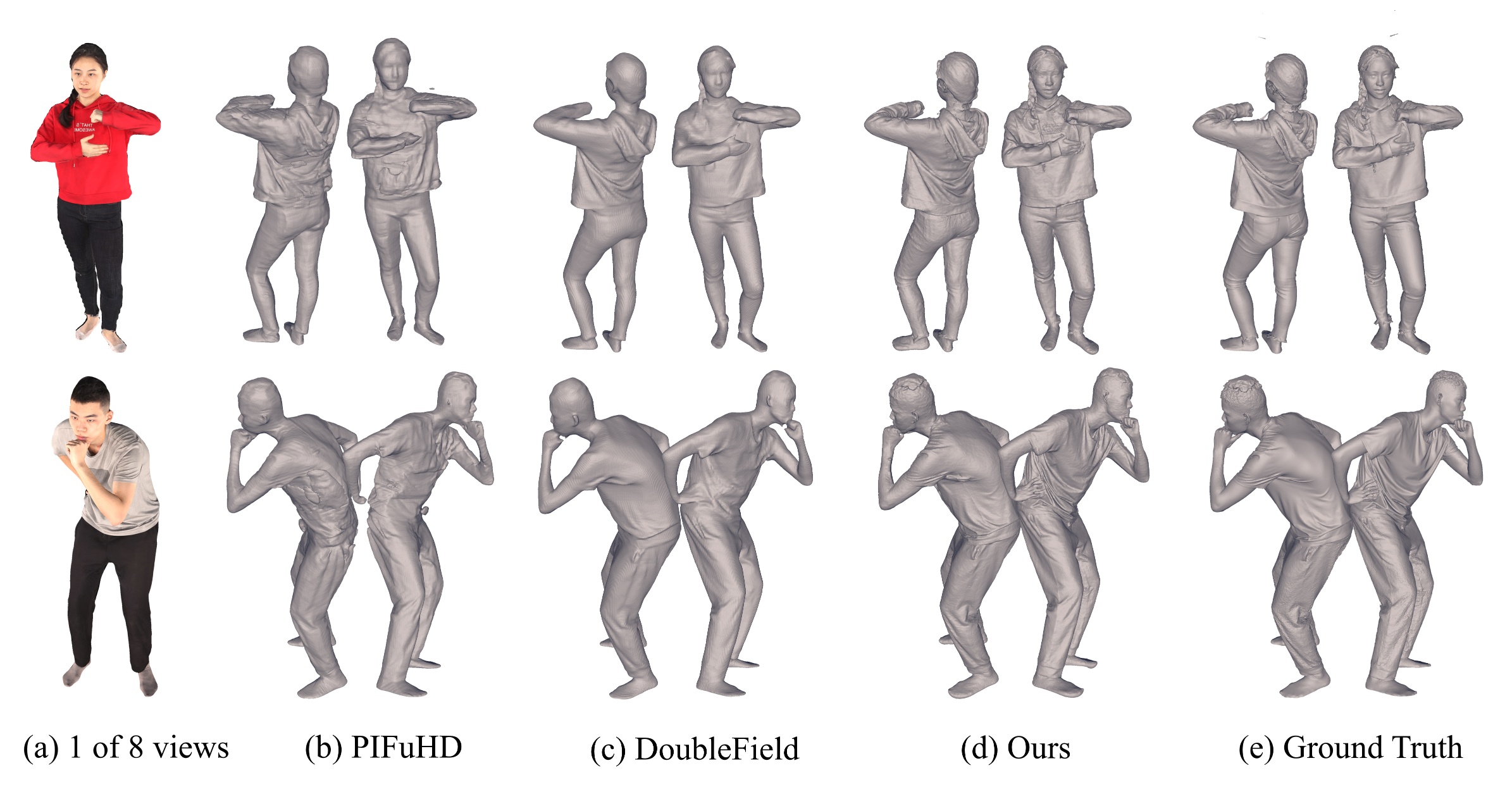}
    \caption{\textbf{Comparison in terms of reconstruction quality.} The human model reconstructed by our method shows significantly higher quality than those reconstructed by the baseline methods. }
    \label{fig:comparison_recon}
\end{figure}

We also compare our method with the state-of-the-art implicit-based human reconstruction methods, including DoubleField~\cite{shao2022doublefield}, PIFuHD~\cite{saito2020pifuhd}, and StereoPIFu~\cite{yang2021stereopifu}. Since PIFuHD focuses on single-view reconstruction and StereoPIFu is mainly proposed for binocular stereo in their original papers, we extend them by adopting the attention mechanism in DeepMultiCap~\cite{zheng2021deepmulticap} to fuse multi-view inputs. We train each method on the same Twindom dataset with the same learning setup. We select 8 views with the view angle of 45° as inputs and quantitatively evaluate the geometry reconstruction using the Chamfer distance and P2S metrics (mm).
To fully demonstrate the high-quality reconstruction performances of our method, we also report the ratio of points in the reconstructed model with different error thresholds (1mm, 2mm, and 5mm).  As shown in Tab.~\ref{tab:recon_table}, our method achieves the best reconstruction quality again. For high-precise reconstruction areas (\ie, within $\leq$ 1mm), our method outperforms the previous methods by more than $42\%$, 
Compared with the coarse human model provided by DoubleField, our method reduces the Chamfer distance and P2S by $54.4\%$ and $44.1\%$, respectively, which proves the efficiency of our conditional stereo network in preserving the color consistency and epipolar constraint under the sparse-view setting.
Moreover, for the reconstruction areas with larger errors ($\leq$ 5mm), our method also has the best performance since the reconstruction of occlusion regions are improved by our light-weight multi-view fusion.
We also show the qualitative results in Fig.~\ref{fig:comparison_recon}, where our method achieves much higher quality of reconstruction in comparison with other methods. 

\begin{figure}[t]
    \centering
    \includegraphics[width=\linewidth]{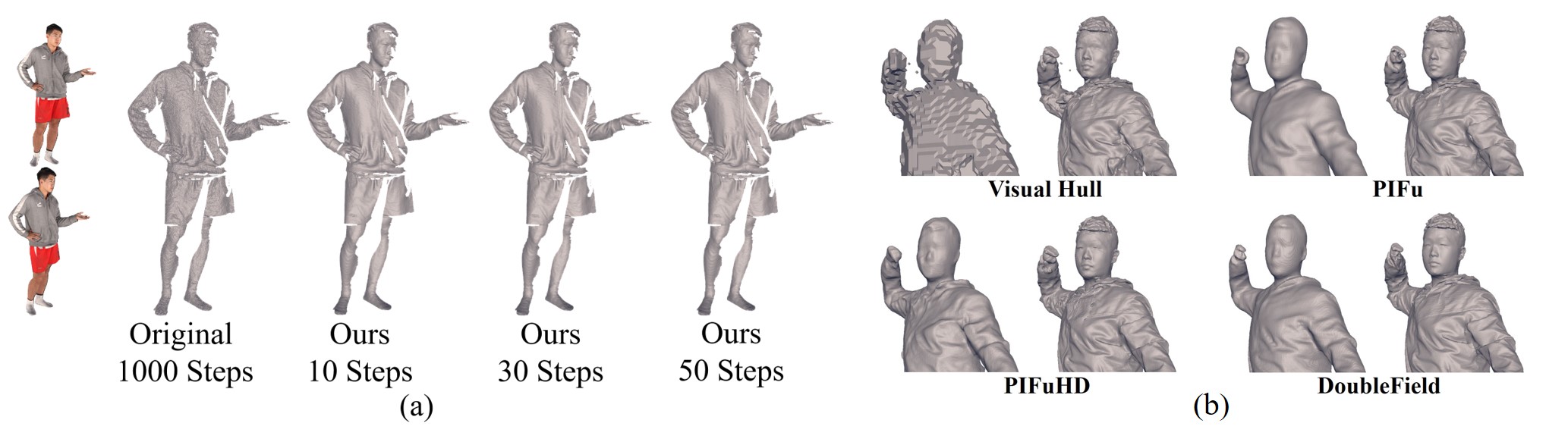}
    \caption{\textbf{(a) Ablation study of diffusion kernel and steps.} The left image shows the unstable noisy result obtained by using the original diffusion kernel in some cases. The right images show the results based on our diffusion kernel in different number of reverse step. \textbf{(b) Results using different coarse model as the initial value.} In each image group, the left shows the coarse model and the right illustrates the diffusion results.} 
    \label{fig:ablation}
\end{figure}
\subsection{Ablation Study}
\noindent \textbf{Different Coarse Model} Our method relies on the coarse human model for diffusion-based stereo. Since the reverse process manually adds noises to the coarse model, our method can handle small errors in the initial depth.
As shown in Fig.~\ref{fig:ablation}(b), even using the visual hull (VH) from 8 views as the initial coarse model, our method can still recover details in visible region.

\noindent \textbf{Diffusion Model} To validate the key components of our multi-level diffusion stereo network, we remove the diffusion model from the stereo network. Such a modified network can be regarded as an iterative refinement model which is similar with Raft-Stereo~\cite{lipson2021raft}. Tab.~\ref{tab:stereo_table} reports the quantitative results and shows that the modified network can only achieve the performances competitive to Raft-Stereo~\cite{lipson2021raft}, which validates that the diffusion model is one of the key components in our diffusion-based stereo.

\noindent \textbf{Diffusion Kernel and Step} As mentioned in Sec.~\ref{sec:diffusion_process}, we adopt a different diffusion kernel for faster and more accurate disparity flow estimation. To validate our new kernel, we report results under the different kernels in Tab.~\ref{tab:stereo_table}. For the method using the original diffusion kernel, we adopt a 1000-steps diffusion model and the same learning setup to train the network. As shown in Tab.~\ref{tab:stereo_table}, the stereo network based on the original diffusion kernel is not as accurate as ours. 
% We think it comes from its potential generative ability. 
In addition, we found that the original diffusion kernel is not stable for stereo matching in our experiments. As shown in Fig.~\ref{fig:ablation}, it generates noisy depth maps in some cases. We also ablate our diffusion model with different steps and illustrate results in Fig.~\ref{fig:ablation}.  Our method is more efficient in the reverse process and can recover details in only 10 steps. For time cost, our method takes about 50ms in one forward step and predicts the depth maps of 8 views in 12s with 30 diffusion steps, which is much faster than the original 1000-step diffusion model.

%% file: section_conclusion.tex
\section{Discussion}
\noindent\textbf{Conclusion.}
We introduced DiffuStereo, a novel system for reconstructing high-quality 3D human models from sparse-view RGB images. With an initial estimate of the human model, our system leverages a novel iterative stereo network based on diffusion models to produce highly accurate depth maps from every two neighboring views. This diffusion-based stereo network is carefully designed to handle sparse-view, high-resolution inputs. The high-quality depth maps can be assembled to generate the final 3D model. Compared to existing methods, ours can reconstruct sharper geometrical details and achieve higher accuracy.  

\noindent\textbf{Limitation.}
The main limitation of our method is the dependence on DoubleField to estimate an initial human model. In addition, our method cannot reconstruct the geometric details in invisible regions due to the lack of observation in our sparse-view setting. 
 
\noindent\textbf{Acknowledgements.} This paper is supported by National Key R\&D Program of China (2021ZD0113501) and the NSFC project No.62125107 and No.61827805.